\documentclass[preprint,12pt]{elsarticle}

\usepackage{amssymb}
\usepackage{subfigure}
\usepackage{setspace}
\usepackage{geometry} 

\usepackage{multirow}
\usepackage{amsmath}
\usepackage{mathrsfs} 
\newtheorem{definition}{Definition}
\usepackage{floatrow}
\newfloatcommand{capbtabbox}{table}[][\FBwidth]
\usepackage[colorlinks,linkcolor=blue]{hyperref}

\journal{ISPRS Journal of Photogrammetry and Remote Sensing}

\begin{document}

\begin{frontmatter}




\title{A Saliency-Guided Street View Image Inpainting Framework for Efficient Last-Meters Wayfinding}


\author[inst1]{Chuanbo Hu}

\affiliation[inst1]{organization={Lane Department of Computer Science and Electrical Engineering, West Virginia University},
            addressline={395 Evansdale Dr}, 
            city={Morgantown},
            postcode={26505}, 
            state={West Virginia},
            country={United States}}

\author[inst2]{Shan Jia\corref{cor1}}
\affiliation[inst2]{organization={University at Buffalo, State University of New York},
            addressline={12 Capen Hall}, 
            city={Buffalo},
            postcode={14260}, 
            state={New York},
            country={United States}}

\author[inst3]{Fan Zhang}
\affiliation[inst3]{organization={Department of Civil and Environmental Engineering, The Hong Kong University of Science and Technology},
            addressline={Clear Water Bay, Kowloon}, 
            city={Hong Kong},
            country={China}}

\author[inst1]{Xin Li}
\cortext[cor1]{Corresponding author (shanjia@buffalo.edu)}

\begin{abstract}
Global Positioning Systems (GPS) have played a crucial role in various navigation applications. Nevertheless, localizing the perfect destination within the last few meters remains an important but unresolved problem. Limited by the GPS positioning accuracy, navigation systems always show users a vicinity of a destination, but not its exact location. Street view images (SVI) in maps as an immersive media technology have served as an aid to provide the physical environment for human last-meters wayfinding. However, due to the large diversity of geographic context and acquisition conditions, the captured SVI always contains various distracting objects (e.g., pedestrians and vehicles), which will distract human visual attention from efficiently finding the destination in the last few meters. 
To address this problem, we highlight the importance of reducing visual distraction in image-based wayfinding by proposing a saliency-guided image inpainting framework. It aims at redirecting human visual attention from distracting objects to destination-related objects for more efficient and accurate wayfinding in the last meters. Specifically, a context-aware distracting object detection method driven by deep salient object detection has been designed to extract distracting objects from three semantic levels in SVI. Then we employ a large-mask inpainting method with fast Fourier convolutions to remove the detected distracting objects. Experimental results with both qualitative and quantitative analysis show that our saliency-guided inpainting method\footnote{Codes are available at \href{https://github.com/cbhu523/saliency\_last\_way\_finding/}{\url{https://github.com/cbhu523/saliency\_last\_way\_finding/}.}} can not only achieve great perceptual quality in street view images but also {\em redirect} the human's visual attention to focus more on static location-related objects than distracting ones. The human-based evaluation also justified the effectiveness of our method in improving the efficiency of locating the target destination.
\end{abstract}


    
     
      



\end{frontmatter}


\section{Introduction}
\label{sec:sample1}
The worldwide coverage of the Global Positioning System (GPS) has made it a standard means for navigation in modern life. Ranging from map applications (e.g., Google Maps) to navigation in mobile platforms, GPS has provided accurate and reliable localization services for billions of users. However, limited by the horizontal accuracy of approximately 5 meters in smartphone-based GPS~\cite{saha2019closing}, even around 10 meters in an urban environment~\cite{merry2019smartphone}, identifying the exact location after being navigated to a destination is still challenging for users, especially in highly urbanized areas (e.g., urban canyons, as shown in Figure \ref{fig:1} (a) and (d)). This challenge is called \textit{the last-meters wayfinding problem} \cite{saha2019closing, shu2015last, lock2017portable}.

Existing studies on last-meters wayfinding focus primarily on improving accuracy for people with visual impairments~\cite{saha2019closing, lock2017portable}. However, for average users, relying on human vision to identify the destination is also not an easy task, such as in strange surroundings or highly diverse and dynamic environments~\cite{shu2015last}. To address this problem, users tend to use Street View Images (SVI) in Google Maps to virtually explore the precise destination in the last few meters before starting~\cite{anguelov2010google, mahabir2020crowdsourcing}. Although SVI, as an interactive panorama from the street, has allowed users to virtually visit remote areas and intuitively query the target destination (such as stores, hotels, gas stations, and parks), there is always confusion in identifying the exact destination in the last few meters. 
The high diversity of geographic context and acquisition conditions of the SVI will inevitably cause visual distraction due to the fact that various large-scale objects in the SVI, such as pedestrians, vehicles, and road signs, will distract human attention from finding the destination (see the example in Figure \ref{fig:1} (b) and (e)).

\textbf{\begin{figure*}[t]
  \centering
  \includegraphics[width=1\linewidth]{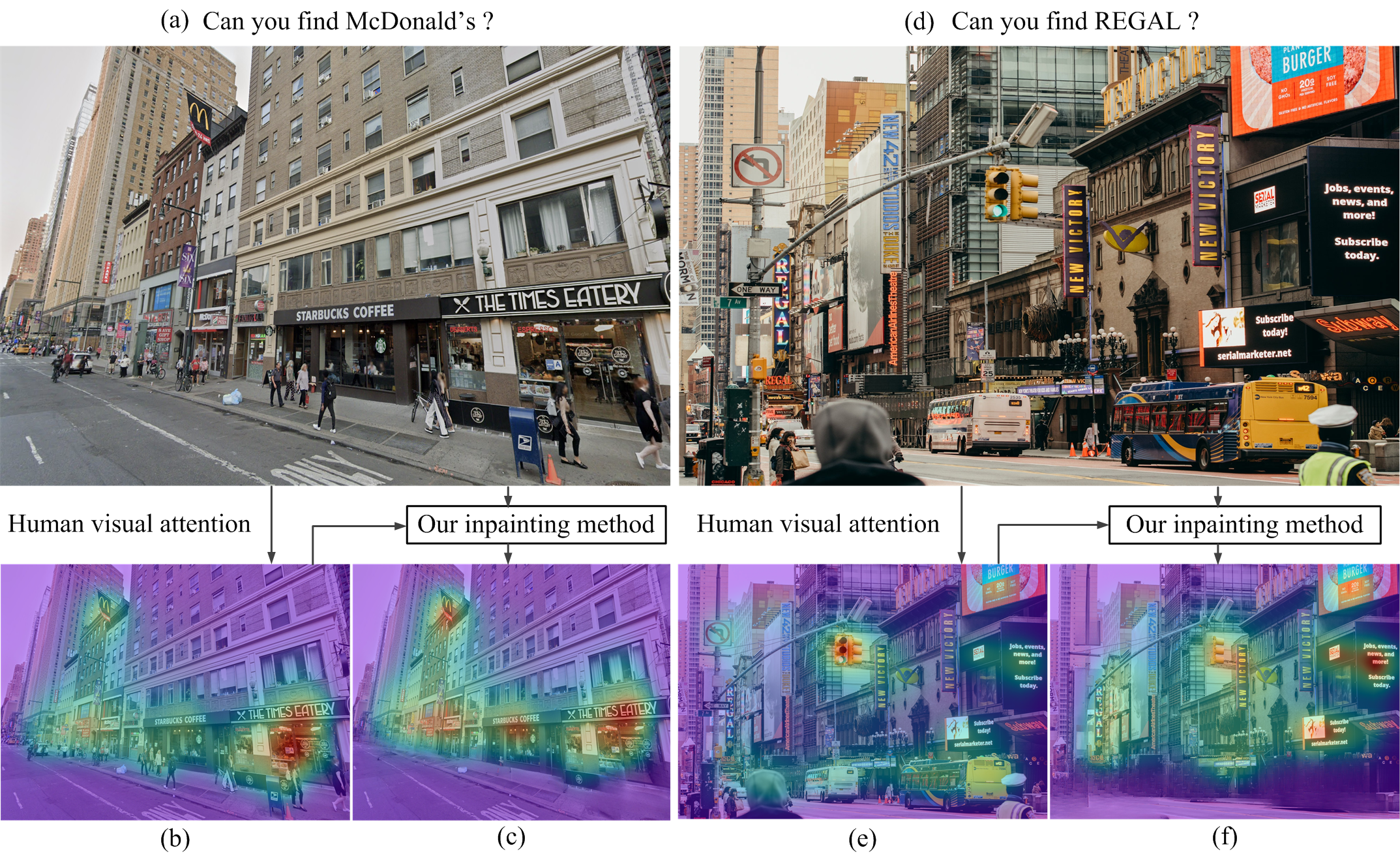}
  \vspace{-0.55cm}
  \caption{The proposed saliency-guided street view image inpainting method can successfully remove distracting objects to redirect human visual attention to static landmarks. Note the shift of salient regions highlighted by the warm color before and after inpainting.}
  \label{fig:1}
\end{figure*}}

    \vspace{-0.55cm}
Unlike existing studies that propose to improve last-meters route finding based on additional signals~\cite{shu2015last, li2020multi} or fine-grained map generation~\cite{wang2018crowdnavi}, we highlight the importance of relying on human vision for last-meters route finding, while proposing to improve location efficiency by reducing visual distraction in street view images. Since there is little research on improving destination finding in SVI using visual attention from humans, how to facilitate the task of SVI-based wayfinding by suppressing those distracting objects remained unexplored.

Humans have a tremendous ability to quickly adjust their gaze when observing a scene to obtain selected visual information \cite{sun2018towards}. This ability enables them to deploy limited processing resources to the most relevant visual information and to understand complex real-world scenes efficiently and accurately. Understanding and simulating this redirecting mechanism has scientific and practical values. One of the most important goals of vision research is to reveal the mechanisms of visual attention and the fixation behavior of human visual systems. Eye tracking, as the technological process of recording gaze movements and viewing patterns, can help researchers reveal the mechanisms of visual attention. With an eye tracking camera, eye movements such as fixations, saccades, and smooth pursuit can be captured \cite{lander2017inferring}. The lengths/durations of these movements vary with the subconscious state of the brain. However, saliency modeling for SVI has remained an underresearched territory in both vision and geospatial communities. 

The mechanism of human visual attention has been explored based on simulations and experiments in several large-scale data sets based on eye tracking (e.g., MIT300, MIT1003 \cite{vig2014large}, SALICON \cite{jiang2015salicon} and DHF1K \cite{wang2018revisiting}). For example, in \cite{xu2014predicting}, the authors demonstrated the importance of object and semantic information in predicting human gaze, especially face and text, which have been shown to outweigh other attributes (e.g., motion, sound, touch, watchability, and operability). However, moving objects in SVIs that attract visual attention often distract noise from the task of finding your way. It is desirable to suppress distraction from moving objects and better support our navigation by redirecting visual attention. 

Therefore, in this paper, we make the first attempt to improve human last-meters wayfinding by reducing visual distractions in image-based navigation. We present a hierarchical strategy for context-aware salient object detection from SVIs and design a saliency-guided inpainting method to redirect visual attention. Specifically, inspired by saliency models developed for visual perception \cite{itti1998model, qin2020u2}, we propose to construct a context-sensitive saliency detection model from three semantic levels: walking pedestrians, moving vehicles, and man-made objects. Such hierarchical decomposition of the saliency map allows us to locate and remove distracting objects at the semantic level instead of the object level based on their attraction to human visual attention. Taking into account the interaction between saliency detection (mask generation) and image inpainting (object removal), we can redirect human visual attention to the task of wayfinding. 

\subsection{Motivations}
Given that various methods have been proposed to improve navigation performance, why is our method necessary? The differences of our study from existing methods which primarily focus on the navigation performance of a camera (e.g., in mobile devices or driving systems) are twofold: i) we aim to improve human vision-based navigation; ii) we only focus on the wayfinding in the last meters that current GPS systems can hardly cover. We justify the motivations by providing two possible applications, as follows.

\begin{itemize}
\item \textit{Navigation based on SVI on maps}. A user who has never been to a destination before often relies on the SVI on maps to explore the exact location before departure. However, street view images may contain various large-scale distracting objects as a result of the high diversity of acquisition conditions. Thus, it is always difficult to find the exact destination, especially in images captured in busy commercial districts or road sections (see McDonald's example in Figure \ref{fig:1} (a) or REGAL in Figure \ref{fig:1} (d)). In this application setup, our work can serve as a post-processing tool for the SVIs before their release in maps. We can extract and remove distracting visual objects based on our saliency-guided image inpainting framework, which will not only redirect human attention to destination-related objects, but can also provide privacy protection for attractive but sensitive objects (as shown in Figure \ref{fig:1} (c) and (f)).
\item \textit{Augmented Reality (AR) navigation}, which provides directions to users on the screen overlaid on top of real environments seen through the camera of a device (e.g., a smartphone or headset). Based on GPS data, such scenarios also occur where a user is brought to the vicinity of a destination (especially in outdoor scenarios) but cannot find the exact location due to many distracting objects in the captured images. In this case, our work can serve as a post-processing tool in AR navigation systems to reduce visual distraction only in the last few meters and redirect human attention to the destination finding.
\end{itemize}

\subsection{Contributions}
The technical contributions\footnote{Codes of the designed framework are available at \href{https://github.com/cbhu523/saliency\_last\_way\_finding/}{\url{https://github.com/cbhu523/saliency\_last\_way\_finding/}.}} of this paper are summarized below.

\begin{itemize}
    \item \textit{Context-aware salient object detection}. {{Unlike existing methods using object-level regions for editing,}} we propose to construct a hierarchical decomposition of the saliency map, which allows us to detect and extract distracting objects at semantic levels in street view images.
    
     \item \textit{Saliency-guided image inpainting to redirect visual attention.} Unlike the previous formulation of the image-inpainting model, we focus on the interaction between saliency detection and image inpainting. The context-aware saliency map facilitates the task of inpainting; in the meantime, the image inpainting achieves the objective of redirecting visual attention to facilitate the task of human wayfinding.
     
     \item {{\textit{Evaluation of visual attention changes.} 
     We design two metrics to quantitatively measure changes in human attention in distracting objects and target regions. For the first time, we experimentally verified the intuition of reducing visual distraction to accelerate the SVI-based wayfinding task. A platform is also designed to compare human search time before and after redirecting visual attention to SVIs.}}
     
      \item {{\textit{Improved efficiency for last-meters wayfinding with good perceptual quality.} Unlike previous methods requiring well-defined or user-provided editing masks as input, the proposed framework takes a single street view image as input and automatically outputs the inpainted image by removing distracting objects. Extensive evaluation demonstrates the benefit of saliency-guided image inpainting to accelerate the task of last-meters wayfinding in street view images and indoor scenarios.}}
      
\end{itemize}

The remainder of the paper is organized as follows. Section 2 provides an overview of related studies. Section 3 describes the proposed saliency-guided street view image inpainting approach. In Section 4, we present the experimental evaluation and results, including both qualitative and quantitative comparisons. We discuss research limitations and potential applications in Section 5, and draw some conclusions about future research directions in Section 6.

\section{Related Work}\label{sec:2}
This study aims to apply a saliency-driven image inpainting technique to reduce human visual distraction in street view images for last-meters wayfinding. In this section, we cover three related areas, including last-meters wayfinding, image-inpainting techniques, and saliency detection methods for human visual attention prediction.

\textbf{Last-meters wayfinding}. The last-meters wayfinding problem refers to the problem of finding the precise geographic position of a destination after being guided to an approximate destination using map data \cite{saha2019closing}. These last meters of navigation cannot be covered by current GPS systems for two main reasons: lack of granular map data and loss of absolute positioning. To bridge the gap between the user's exact destination and the end position provided by current navigation services, several studies have been proposed in recent years. A category of existing methods explored signal/sensor-assisted approaches for highly accurate location, including WiFi signal strength~\cite{turner2011empirical, yang2012locating, roy2021survey} for indoor localization, audio information~\cite{mcgookin2009audio} for tourist wayfinding, and geomagnetic field~\cite{shu2015last} for indoor and outdoor navigation. Another category of solutions proposes the use of existing techniques to improve last-meters wayfinding, such as crowd-sourcing techniques~\cite{shen2013walkie, wang2018crowdnavi} and AI methods~\cite{saha2019closing} to capture the relationship between landmarks and the person's behavior information for people with visual impairments.

Despite these advances in improving the efficiency and accuracy of navigation systems for human last-meters wayfinding, the need for additional and precise signal extraction or a sustainable incentive crowd-sourcing mechanism limits its applicability. To provide a simpler, more applicable, and effective solution, we emphasize the importance of improving human visual attention for image-based last-meters wayfinding considering that street view images have been widely used in navigation systems. Due to the process of gathering SVI in the public, it inevitably captures diverse information, including static objects (e.g., landmarks and roads) and dynamic objects (e.g., vehicles, pedestrians, and traffic signs). However, dynamic objects may not only cause occlusion both visually and pragmatically, but also collide with privacy issues with identity-related information in many SVI applications, especially the most widely used image-based localization. Therefore, several studies have proposed to clean SVIs by blurring sensitive information~\cite{sebastian2019lidar}, replacing completely anonymous pedestrians~\cite{nodari2012digital}, replacing sensitive objects with synthetic content generated by generative adversarial networks (GAN)~\cite{yu2021gan}, or removing dynamic objects using image-inpainting techniques~\cite{uittenbogaard2019privacy}. Among them, advanced image-inpainting methods have shown superiority in removing dynamic objects with high visual quality.

\textbf{Image inpainting}. Image inpainting refers to a class of techniques that can remove unwanted objects from images. The removal of pedestrians from the SVI was first studied \cite{flores2010removing} from a privacy point of view, where pedestrian detection and pixel cloning methods were used. Rapid advances in deep learning in recent years have provided new weaponry for powerful inpainting techniques, e.g., contextual attention \cite{yu2018generative}, partial convolutions \cite{liu2018image}, gated convolution \cite{yu2019free}, and recurrent feature reasoning \cite{li2020recurrent}. Most recently, the knowledge distillation strategy was adapted to the problem of image inpainting \cite{suin2021distillation}; a novel dynamic selection network (DSNet) was proposed to distinguish the corrupted region from the valid ones throughout the entire network architecture. In \cite{suvorov2022resolution}, robust Large Mask (LaMa) inpainting with Fourier convolutions was developed with resolution, which will serve as the baseline for our research. Note that almost all existing image-inpainting methods assume that the inpainting domain (mask) is random or known as a priori.

\textbf{Saliency detection}. Saliency detection \cite{itti1998model} is a problem that has been extensively studied in the literature on visual perception and computer vision. Various saliency models, including context-aware \cite{goferman2011context}, spectral residual \cite{hou2007saliency}, and hierarchical \cite{yan2013hierarchical}, have been proposed in the literature. Among them, hierarchical saliency detection \cite{yan2013hierarchical} shares a motivation similar to ours. However, a significant departure from the previous work \cite{yan2013hierarchical} is the definition of hierarchy - in the context of last-meters wayfinding, we will consider a semantically meaningful definition of hierarchy in this work. More recently, deep learning for saliency detection has also been extensively studied - from the unification of global context with local context \cite{zhao2015saliency} to the Holistically Nested Edge Detector (HED) \cite{hou2007saliency}. For the current state-of-the-art in deep-saline models, refer to recent reviews such as \cite{cong2018review} and \cite{zhang2018review}. 

{Furthermore, there have been some recent studies proposed to combine the saliency with image editing techniques~\cite{chen2019guide, mechrez2019saliency, mejjati2020look, jiang2021saliency, aberman2021deep}. However, these methods cannot be applied to the street-view image based wayfinding for the following three reasons.
1) The previous methods use recoloring to emphasize the target objects or attenuate the distracted regions~\cite{chen2019guide, mechrez2019saliency, mejjati2020look, aberman2021deep}. They not only require accurate masks, but also cannot handle the privacy disclosure in street view images with pedestrians or vehicles. 
2) They all require the input of a well-defined or user-provided mask as the manipulation region~\cite{chen2019guide, mechrez2019saliency, mejjati2020look, jiang2021saliency, aberman2021deep} and some can only handle a single or finite-size region~\cite{jiang2021saliency, mejjati2020look}.
3) In terms of the evaluation of attention retargeting, these methods compare the similarity of ground-truth masks/saliency and saliency maps of manipulated images~\cite{chen2019guide, mechrez2019saliency, mejjati2020look, jiang2021saliency}, which cannot fully represent changes in human attention in both distracting regions and target regions.}

\textbf{New insight}. Street-view image applications face two obstacles in the real world: {\em occlusion} and {\em privacy}. The former is related to the application of last-meters wayfinding in that the region of interest might be occluded by distracting objects, which makes it more challenging to find the match. The latter generally refers to the protection of sensitive information (e.g., biometrics and vehicle license plates) in the public domain. Saliency-guided image inpainting seems to be a promising technical solution to solve both problems. However, to the best of our knowledge, saliency detection and image inpainting have been studied separately in the open literature. Their rich interaction can be justified by 1) the detected salient objects serving as the inpainting domain; and 2) removing distracting objects by inpainting could redirect visual attention to landmarks such as static buildings in street view images. How to jointly exploit the interaction between saliency detection and image inpainting remains an unexplored research topic in street view images. 

\section{Methodology}\label{sec:3}

\begin{figure}[!t]
  \centering
  \includegraphics[width=1\linewidth]{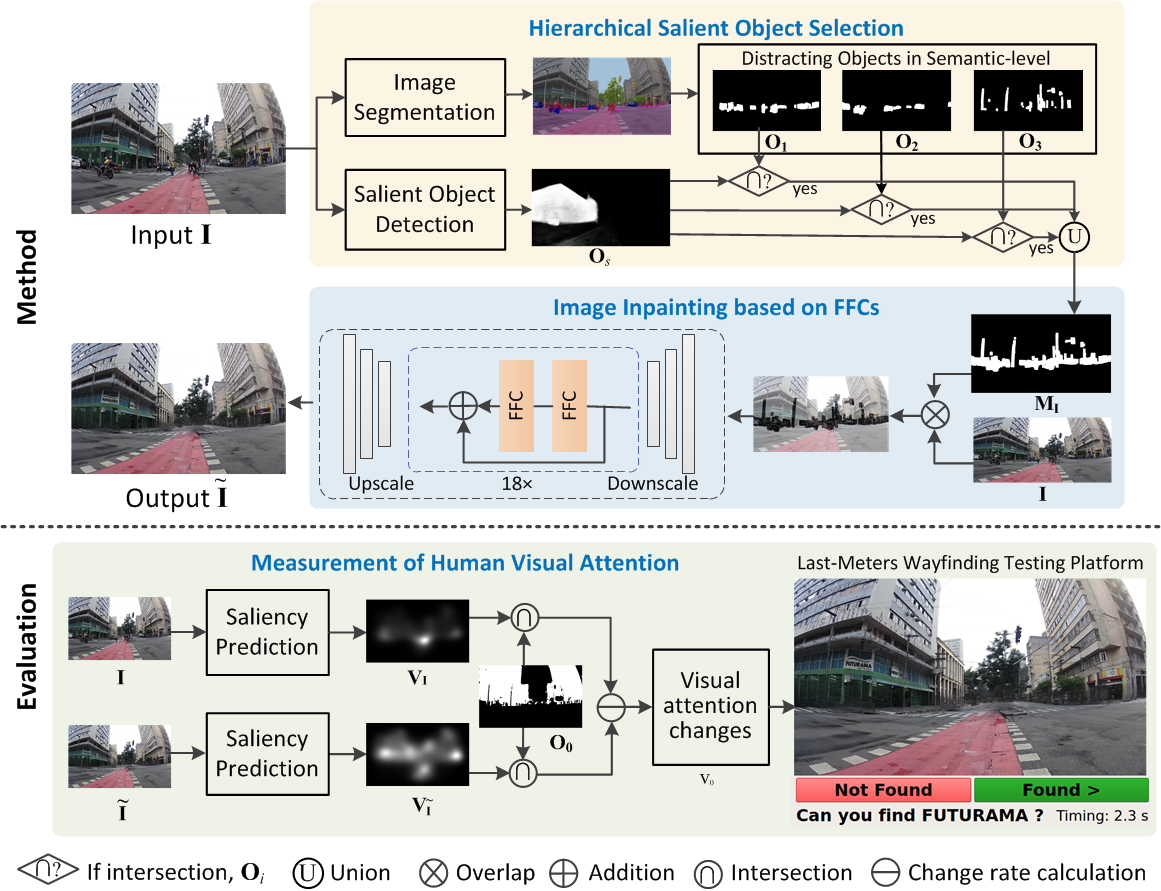}
  \caption{Overview of the proposed saliency-guided street view image inpainting framework. It consists of three building blocks: hierarchical salient object selection, saliency-guided image inpainting based on fast Fourier convolutions (FFCs), and measurement of human visual attention by visual attention changes and a self-developed last-meters wayfinding testing platform. Note that modeling the interaction between saliency detection and image inpainting leads to effective removal of distracting objects for last-meters wayfinding.}
  \label{fig:2}
\end{figure}

In this study, we propose a saliency-guided street-view image-inpainting framework to remove distracting objects for efficient last-meters wayfinding. We first formulate the problems in terms of last-meters wayfinding and SVI inpainting in this section. Then we introduce the details of the proposed framework with the following three components: 1) hierarchical salient object detection; 2) image inpainting based on Fourier convolution for distracting object removal; 3) measurement of human visual attention for last-meters wayfinding. The overview of the proposed saliency-guided inpainting framework is illustrated in Figure \ref{fig:2}.

\subsection{Problem Formulation}\label{sec:3.1}
Given a street view image of an approximate destination guided by map data, the goal of human last meters wayfinding in our study is to reduce visual distraction in the image and guide human vision to efficiently locate the precise geographic location. 
Let the street view image be $\textbf{I}$, which could contain various landmarks and distracting objects with different levels of importance for last-meters wayfinding. Semantic image segmentation techniques~\cite{liu2019recent} can be employed to partition the image into multiple segments (sets of pixels, also known as image objects), and denoted $\textbf O_i\in \textbf I \;(1\le i\le N, N$ is the number of object categories). Based on the importance level of the objects in street view images for last-meters wayfinding, we give the following definitions to classify the objects into three categories. 

\begin{definition}[Object of interest]
An object of interest for last-meters wayfinding is the object that is location-related. It generally contains static buildings, such as landmarks, shopping malls, stores, and supermarkets.
\end{definition}
\vspace{-0.25 cm}
\begin{definition}[Distracting object]
A distracting object for last-meters wayfinding is the object that distracts a human's visual attention from searching and identifying the location-related landmarks in the street view image. It usually contains moving vehicles, pedestrians, and dynamic traffic signs. 
\end{definition}
\vspace{-0.25 cm}

\begin{definition}[Neutral object]
A neutral object for last-meters wayfinding is the object that is neither location-related nor eye-catching in the street view image. It usually contains a sky and vegetation.
\end{definition}

Given the above definitions, we formally define the problem of street view image inpainting in last-meters wayfinding. Let $\textbf O_o, \textbf O_d$ be objects of interest and distracting objects, respectively, and the goal of street view image inpainting is to inpaint the distracting objects $\textbf{O}_d$ in the images and redirect visual attention to objects of interest $\textbf{O}_o$, shown below. 
\begin{equation}
\mathcal M(\textbf{I})=\textbf{O}_d \xrightarrow{\widetilde{\textbf{I}} = \mathcal I(\textbf{I})}\mathcal M(\widetilde{\textbf{I}})=\textbf{O}_o 
\vspace{-0.15cm}
\end{equation}
where $\mathcal M(.)$ is the visual attention prediction function (gaze), $\mathcal I(.)$ is the image inpainting operation, and $\widetilde{\textbf{I}}$ is the image inpainted. To achieve this goal, we will first propose an approach to detect the existence of distracting objects $\textbf{O}_d$ in street view images. Note that these distracting objects $\textbf{O}_d$ are extracted from different semantic levels based on the image content. Then we will design an image inpainting method $\mathcal I(.)$ to remove distracting objects for the reduction of visual distraction and finally introduce the saliency prediction $\mathcal M(.)$ to predict and measure human visual attention. To facilitate our presentation, Table \ref{tab:0} summarizes the mathematical symbols used in this section.
\begin{table}[t]
\centering
\caption{List of symbols used to describe the proposed method and what they mean.}
\vspace{0.2cm}
\begin{tabular}{c|c} 
\hline
\textbf{Notation}  & \textbf{Meaning}      \\ 
\hline
$\textbf{I}$  & Input street view image               \\ 
\hline
$\widetilde{\textbf{I}}$ & Inpainted street view image           \\ 
\hline
$\textbf{O}_o$ & Object region of interest    \\ 
\hline
$\textbf{O}_d$  & Distracting object region \\ 
\hline
$\textbf{O}_0$  & Object region in Semantic level 0 (Building)    \\ 
\hline
$\textbf{O}_1$  & Object region in Semantic level 1 (Human)         \\ 
\hline
$\textbf{O}_2$  & Object region in Semantic level 2 (Vehicle)       \\ 
\hline
$\textbf{O}_3$  & Object region in Semantic level 3 (Sign)            \\ 
\hline
$\textbf{O}_s$ & Salient object region\\ 
\hline
$\textbf{M}_\textbf{I}$ & Inpainting mask  \\ 
\hline
$\textbf{V}_\textbf{I}$ & Visual attention map of $\textbf{I}$  \\ 
\hline
$\textbf{V}_{\widetilde{\textbf{I}}}$ & Visual attention map of  $\widetilde{\textbf{I}}$  \\ 
\hline
$\textbf{X}, \textbf{X}^l, \textbf{X}^g$ & Input of FFC module \\ 
\hline
$\textbf{Y}, \textbf{Y}^l, \textbf{Y}^g$ & Output of FFC module\\ 
\hline
$f$ & Convolution operations of FFC\\ 
\hline
$\gamma$ & Threshold to binarize the object saliency map     \\ 
\hline
$\lambda$ & Loss function weights  \\ 
\hline
$v_o$ & Visual attention changes on the object of interest \\ 
\hline
$v_d$ & Visual attention changes on distracting object\\ 
\hline
$\mathcal L$ & Loss function  \\ 
\hline
$\mathcal I(.)$  & Image inpainting function             \\ 
\hline
$\mathcal M(.)$    & Visual attention prediction function  \\
\hline
\end{tabular}
\label{tab:0}
\end{table}
\subsection{Hierarchical Salient Object Detection}\label{sec:3.2}

Unlike existing image inpainting methods~\cite{yu2018generative, liu2018image, yu2019free, li2020recurrent, suvorov2022resolution, cao2021learning} which are mainly based on randomly or manually selected inpainting masks, we develop an automatic inpainting domain extraction method by detecting context-sensitive salient objects in street view images. 
\begin{figure}[t]
  \centering
  \includegraphics[width=0.8\linewidth]{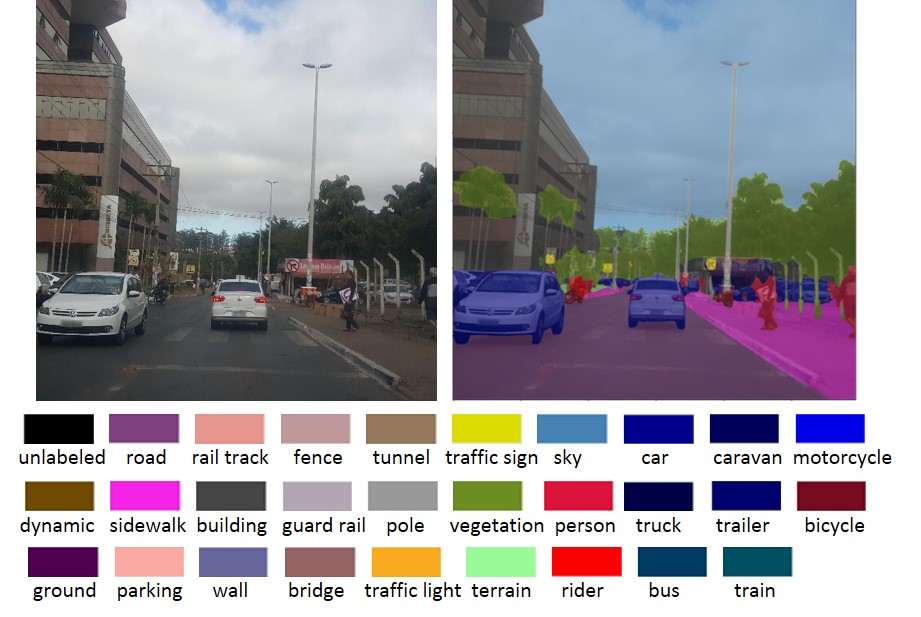}
  \vspace{-0.35cm}
  \caption{Semantic segmentation results and labels on the Cityscapes dataset using DeepLabv3+. Best view in color.}
  \label{fig:2_a}
\end{figure}

\begin{figure*}[t]
  \centering
  \includegraphics[width=1\linewidth]{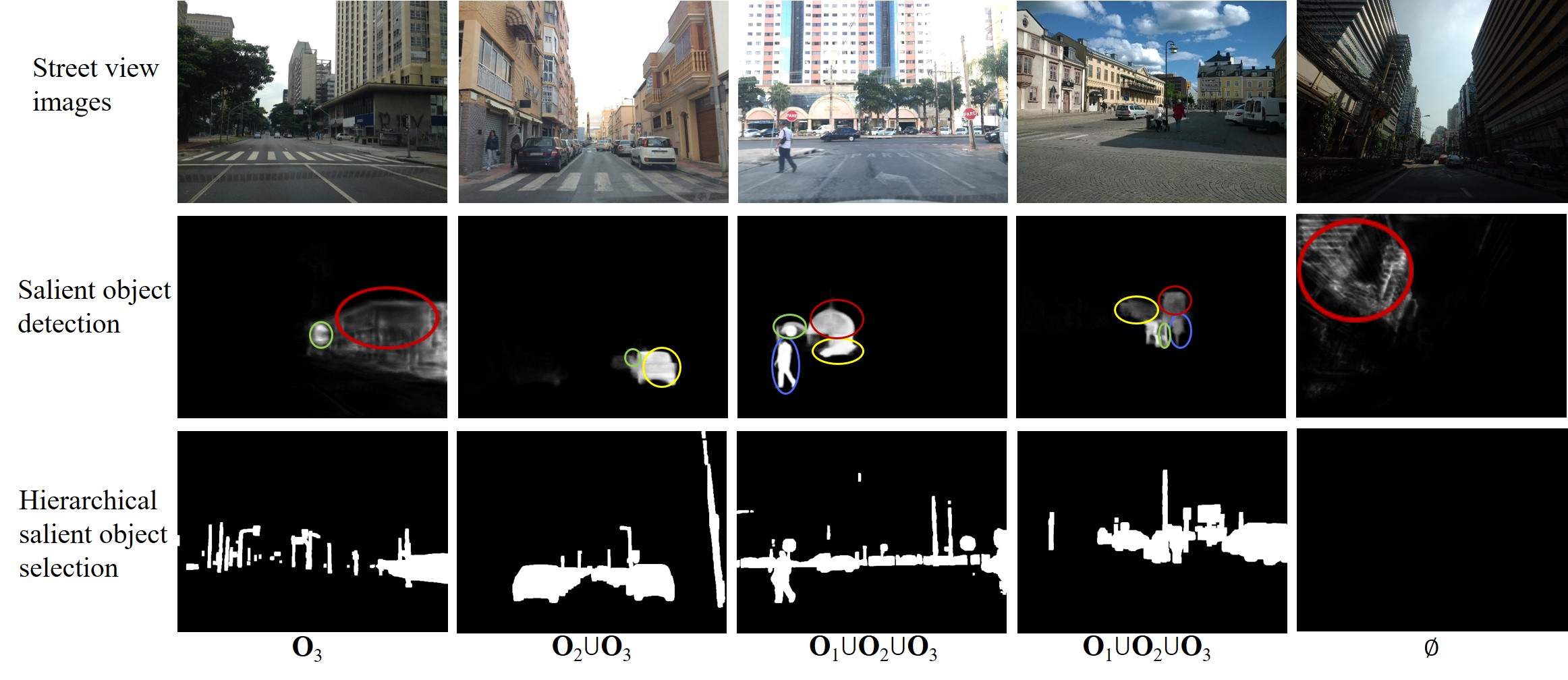}
  \vspace{-0.85cm}
  \caption{Examples of salient object detection and selection in street view images. Red ovals - salient objects in $\textbf O_0$ semantic level (Building), blue - salient objects in $\textbf O_1$ (Human), yellow - salient objects in $\textbf O_2$ (Vehicle), and green - salient objects in $\textbf O_3$ level (Sign). Best view in color.}
  \label{fig:2b}
\end{figure*}

Several previous studies~\cite{borji2012boosting,nguyen2013static,xu2014predicting,he2019understanding} have observed that objects that convey more information, such as humans, faces, cars, text, and animals, tend to be more prominent and guide the human gaze in a scene. However, in street-view images, these salient objects carry no/less location-related information but will distract visual attention from last-meters wayfinding. To remove these distracting objects, we first classify them into three levels and then design a hierarchical distracting object extraction method based on semantic segmentation and salient object detection. Specifically, this process consists of the following steps:

\textbf{1) Image segmentation.} We employ the state-of-the-art semantic image segmentation method, DeepLabv3+~\cite{chen2018encoder} to assign semantic labels to every pixel in a street view image. Due to the encoder-decoder structure with atrous separable convolution, DeepLabv3+ achieves promising segmentation performance, especially on the Cityscape image dataset~\cite{cordts2016cityscapes} with diverse street scenes. Given a street view image as input, the DeepLabv3+ method will capture and output the object boundaries with different semantic labels. The rich semantic information in street view images has been finely classified and labeled into 29 classes in the Cityscapes dataset, as shown in Figure \ref{fig:2_a}. Therefore, the input street view image will be partitioned into several segments as $\textbf O_i$ at the semantic level.

\textbf{2) Salient object detection.}
Image segmentation alone cannot determine distracting objects in street view images. To detect visually attractive objects in images, we then use a deep network with a nested U-structure for the detection of salient objects. Inspired by the powerful multiscale feature extraction of residual blocks, we take a street view image $\textbf{I}$ as input to the $U^2Net$ model \cite{qin2020u2} with 11 stages of residual U blocks and obtain the saliency map with the same spatial resolution as $\textbf{I}$. Each pixel on the map is within the range of [0, 255], and the pixels of the salient object $\textbf O_s$ will have higher values (close to 255). From the first two rows of Figure \ref{fig:2b}, we can observe that informative objects in street view images, such as vehicles, pedestrians, traffic signs, and landmarks, tend to be more attractive as the predicted salient object, which is consistent with previous studies. 
\begin{table}
\footnotesize
\renewcommand{\arraystretch}{1.2}
\newcommand{\tabincell}[2]{\begin{tabular}{@{}#1@{}}#2\end{tabular}} 
\centering
\caption{Four semantic levels for the selection of hierarchical salient objects.}
\vspace{-0.2cm}
\begin{tabular}{c|c|c} 
\hline
\textbf{Category}    & \textbf{Segment} &\textbf{ Semantic level (Fine class)}                \\ 
\hline
\tabincell{l} {Object of interest} & $\textbf O_0$            & Building               \\ 
\hline
\multirow{3}{*}{\tabincell{l} {Distracting object}}& $\textbf O_1$   & Human (Person, Rider)         \\ 
\cline{2-3}
 &$\textbf O_2$           & Vehicle (Car, Truck, Bus, Train, Motorcycle, Bicycle) \\ 
\cline{2-3}
 &$\textbf O_3$       & Sign (Traffic light, Traffic sign, Fence, Pole)     \\
\hline
\end{tabular}
\label{tab:1a}
\end{table}

\textbf{3) Hierarchical salient object selection.}
If we directly take the salient object map after removing the objects of interest as the inpainting mask, one obvious problem is that other objects with the same or similar semantic labels in the image may still catch visual attention. In other words, the salient object map generated by $U^2Net$ is at the instance level, rather than at the semantic level. To address this problem, we designed the hierarchical salient object selection method to extend the salient map from the instance level to the semantic level. In street view images, we mainly focus on the object of interest and distracting objects for last-meters wayfinding. We first classify them into four categories based on the correlation in the semantic labels (see Table \ref{tab:1a}). Next, we use an empirical threshold $\gamma$ to binarize the saliency map, as in previous work~\cite{perazzi2012saliency,yu2018co,wang2020detecting}. Consequently, the salient object region $\textbf O_s$ can be modified to the semantic level as follows.
\begin{equation}
\begin{aligned}
\textbf O_s = \textbf O_1 \; (\text{if} \; \textbf O_s \cap \textbf O_1 \neq \o \;) \\
\cup \; {\textbf O_2} \; (\text{if} \; \textbf O_s \cap \textbf O_2 \neq \o \;) \\
\cup \; {\textbf O_3} \;  (\text{if} \; \textbf O_s \cap \textbf O_3 \neq \o \;) 
\vspace{-0.15cm}
\end{aligned}
\end{equation}
Then we can obtain the final binary inpainting mask $\textbf{M}_\textbf{I}$ with the pixels in $\textbf O_s$ as 1 and the others as 0. Figure \ref{fig:2b} compares the saliency maps before and after the selection of hierarchical salient objects. We can see that more distracting objects can be covered with hierarchical salient object selection as the inpainting mask. 

\subsection{Image Inpainting based on Fast Fourier Convolutions}\label{sec:3.3}

Removing distracting objects at the semantic level means a larger inpainting domain than removing objects at the instance level. Fast Fourier convolutions (FFCs)~\cite{chi2020fast} allow a large receptive field to cover an entire image even in the early layers of the network, and therefore can provide a high perceptual quality and parameter efficiency of the network. Inspired by the recent success of FFCs in the large mask inpainting model LaMa~\cite{suvorov2022resolution}, we build a similar inpainting network for street view images based on the hierarchical salient distracting mask generated.

\begin{figure}[t]
  \centering
  \includegraphics[width=0.9\linewidth]{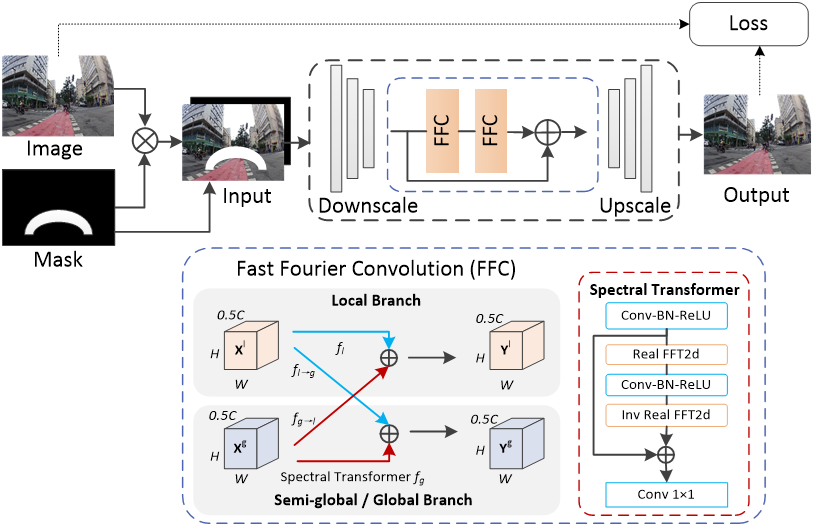}
  \caption{{Overview of the fast Fourier convolutions-based image inpainting model.}}
  \label{fig:2a}
  \vspace{-0.08cm}
\end{figure}
As shown in Figure \ref{fig:2a}, taking the original image $\textbf{I}$ stacked with the binary inpainting mask $\textbf{M}_\textbf{I}$ as input, the inpainting network first downscales the input with three downsampling blocks and then uses 18 residual FFC blocks as generator, followed by three upsampling blocks. Specifically, the FFCs split the input feature map into a local branch with conventional convolutions and a global branch with a real fast Fourier transform (FFT). Let $\textbf X=\left\{ \textbf X^l, \textbf X^g\right\}$ be the input to the FFC module ($\textbf X\in \textbf{R}^{H\times W \times C}$, $\textbf X^l$ be the local part and $\textbf X^g$ the global part), and the output of FFC $\textbf Y=\left\{ \textbf Y^l, \textbf Y^g\right\}$ can be described by the following formulas:
\begin{equation}
\textbf Y^l = \textbf Y^{l\rightarrow l}+ \textbf Y^{g\rightarrow l} = f_l(\textbf X^l) + f_{g\rightarrow l}(\textbf X^g) 
\vspace{-0.15cm}
\end{equation}
\begin{equation}
\textbf Y^g = \textbf Y^{g\rightarrow g} + \textbf Y^{l\rightarrow g} = f_g(\textbf X^g) + f_{l\rightarrow g}(\textbf X^l) 
\vspace{-0.05cm}
\end{equation}
where the component $\textbf Y^{l\rightarrow l} $ is obtained by applying a regular convolution $f_l$ to $\textbf X^l$, $\textbf Y^{g\rightarrow g} $ is obtained by applying a spectral transformer $f_g$ to $\textbf X^g$, and the other two components $\textbf Y^{g\rightarrow l}$ and $\textbf Y^{l\rightarrow g}$ are obtained via an inter-path transition based on regular convolutions to take full advantage of multiscale receptive fields. Specifically, to capture the global/semi-global receptive field in the image, the spectral transform $f_g$ first transforms the original spatial features into the spectral domain, then performs an efficient global update on the spectral data, and finally converts the data back to the spatial format. Real FFT and inverse real FFT are applied to ensure that both input and output are real-valued.

Training in the large-scale Places~\cite{zhou2017places} dataset with a high diversity of scene categories and high-quality photographs, the original LaMa model achieves excellent performance in real-world image editing scenarios. Since street view images for wayfinding are more likely to contain richer information (with complex texture and content) and more diverse image quality (under different light conditions), we follow the pipeline in Figure \ref{fig:2a} to fine-tune the whole inpainting network in street view images. The loss function to train the model is calculated as follows:
\begin{equation}
\mathcal L = \lambda_1 \mathcal L_{adv} + \lambda_2 \mathcal L_{pl1}+ \lambda_3 \mathcal L_{pl2}+ \lambda_4 R_1
\vspace{-0.15cm}
\end{equation}
$\mathcal L_{adv}$ is the adversarial loss to ensure that the inpainting model generates natural-looking local details based on a discriminator network. Perceptual loss $\mathcal L_{pl1}$ is for the high-receptive field, obtained by sequential two-stage mean operations (interlayer mean or intralayer mean) of Fourier convoluted images $\textbf{I}$ and $\widetilde{\textbf{I}}$. Perceptual loss $\mathcal L_{pl2}$ is the loss of feature matching to stabilize training~\cite{wang2018high}, and the regularization term $R_1$ is the discriminator gradient penalty~\cite{suvorov2022resolution} to penalize the discriminator for deviating from Nash equilibrium. $\lambda _1, \lambda _2, \lambda _3$, and $\lambda_4$ are hyperparameters.

\subsection{Measurement of Human Visual Attention for Last-few-meters Wayfinding}\label{sec:3.4}
The salient object detection method, such as the $U^2Net$ model introduced in Section 3.2, is object-level, which cannot accurately represent human visual attention. To predict and measure how the human gaze will change in street view images, we propose to model pixel-level saliency based on the UNISAL network~\cite{drostejiao2020}. 
UNISAL integrates novel domain adaptation modules into an encoder-RNN-decoder-style network. These domain-adaptive modules are designed to shift domains between different saliency datasets and to enable high-quality shared features. 

The entire network was trained in the SALICON image dataset~\cite{jiang2015salicon} and three video datasets with eye-tracking or fixation annotations. Note that the SALICON dataset collected 20,000 natural images with mouse tracking labeling through the Amazon Mechanical Turk (AMT) platform to obtain the ground truth of human gaze positions. Joint training in different modalities results in improved performance of the UNISAL model in visual saliency modeling. To measure human visual attention influenced by image inpainting, we first input the street view images $\textbf{I}$ and the inpainted image $\widetilde{\textbf{I}}$ into the pre-trained UNISAL model, and obtain the predicted visual attention map $\textbf{V}_\textbf{I}$ and $\textbf{V}_{\widetilde{\textbf{I}}}$. Next, we propose to quantitatively represent the increase in visual attention in the object of interest $v_o$ and the reduction of visual attention in distracting objects $v_d$ by
\begin{equation}
v_o = \sum_{i\in \textbf{O}_o}(\textbf{V}_{\widetilde {\textbf{I}}}(i))-\textbf{V}_\textbf{I}(i))/\sum_{i\in \textbf{O}_o}(\textbf{V}_\textbf{I}(i)))
\vspace{-0.15cm}
\end{equation}

\begin{equation}
v_d = \sum_{i\in \textbf{O}_d}(\textbf{V}_\textbf{I}(i) - \textbf{V}_{\widetilde {\textbf{I}}}(i)))/\sum_{i\in \textbf{O}_d}(\textbf{V}_\textbf{I}(i)))
\vspace{-0.15cm}
\end{equation}

\section{Experiments}\label{sec:4}
To evaluate our method, we performed a series of experiments on street view images with quantitative and qualitative analyses. This section first introduces the implementation details and evaluation metrics and then compares the image-inpainting performance with existing methods. Next, we demonstrate the influence of mask selection and city morphology type on inpainting results and changes in visual attention. To show the effectiveness in improving the last-meters wayfinding efficiency, we also conduct human-based evaluation. Finally, we show the applicability of the proposed framework by extending it to indoor scenario applications.

\begin{figure}[!h]
 \centering
 \includegraphics[width=1\linewidth]{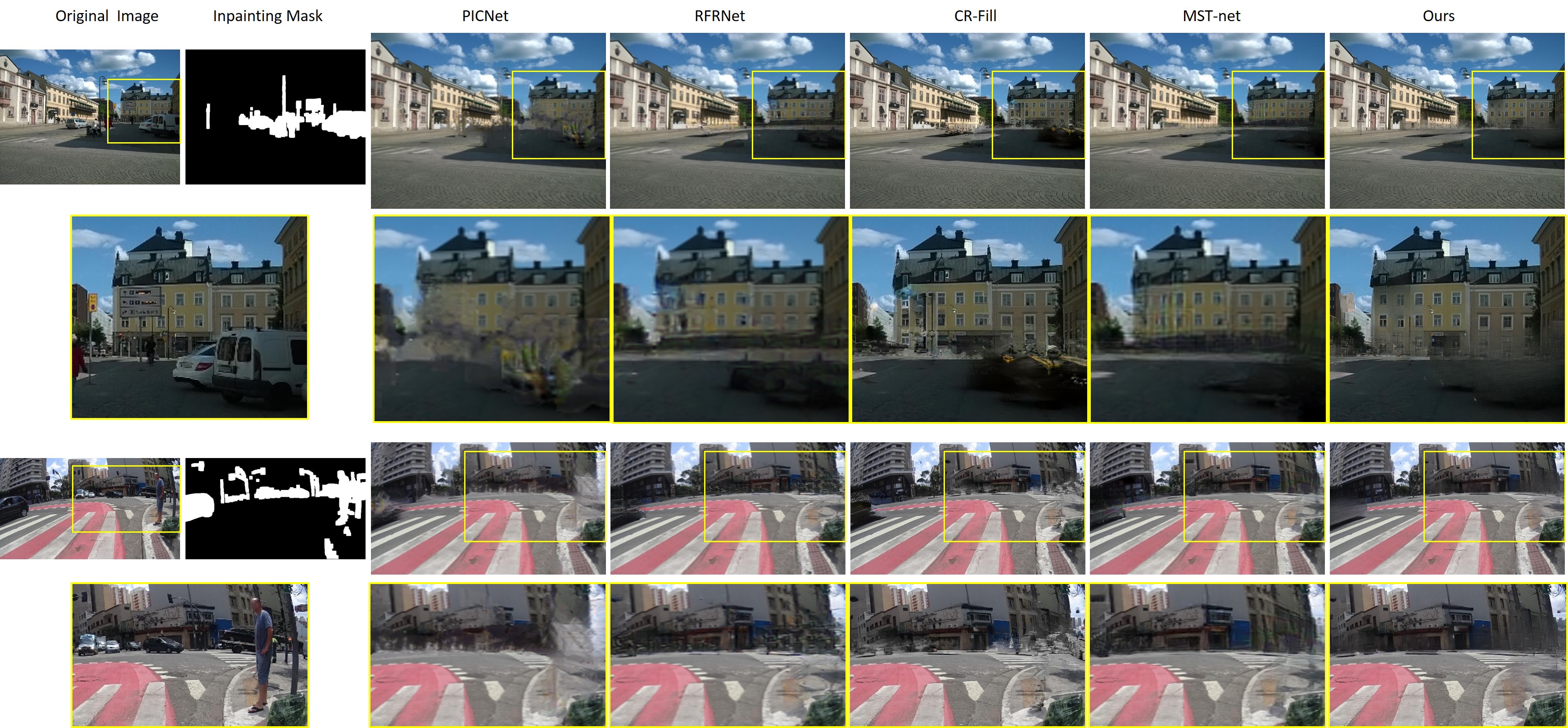}
 \caption{Visual performance of different image inpainting methods on Cityscapes dataset. {Note that all images are adjusted to the same size for a clear comparison.} Best view in color and zoom in.}
  \label{fig:3}
\end{figure}

\subsection{Experimental Settings}\label{sec:4.1}
\textbf{Dataset and implementation details.} We used the large-scale Cityscape dataset~\cite{cordts2016cityscapes} with various street scenes in our experiment. Cityscapes dataset contains a diverse set of stereo video sequences recorded from 50 different cities. For image inpainting, we fine-tuned the big LaMa Fourier model with the training data (18,000 images) in Cityscapes following the settings in the paper~\cite{suvorov2022resolution}, and report all results on the testing subset (5,000 images). Loss weight $\lambda _1$ is set as 10, $\lambda _2$ as 30, $\lambda _3$ as 100, and $\lambda _4$ as 0.001. The threshold $\gamma$ to binarize the object saliency map is set to 0.8, meaning that the pixels larger than the $0.8\times$maximum will be set as 1 and others as 0. For salient object detection, we used the pretrained model of $U^2Net$ model in \cite{qin2020u2}. All experiments were carried out on NVIDIA TITAN RTX GPUs.

\textbf{Evaluation metrics.} We follow the metrics in the recent image inpainting literature~\cite{zheng2019pluralistic,li2020recurrent,yi2020contextual,suvorov2022resolution,zeng2021cr} and report the quantitative results of the mean $l_1$ error, the structural similarity index (SSIM), the peak signal-to-noise ratio (PSNR), the learned perceptual image patch similarity (LPIPS)~\cite{zhang2018unreasonable} and the Frechet Inception Distance (FID)~\cite{heusel2017gans} based on InceptionNet~\cite{szegedy2016rethinking} features. {Since the calculation of $l_1$, SSIM, and PSNR metrics requires the comparison of the inpainted image with ground-truth data, we inserted random objects (including vehicles, pedestrians and traffic signs) into the image in the Cityscapes dataset~\cite{imam2019semantic}. The processed images were input to different inpainting models with the inserted object region as the inpainting mask. Correspondingly, the $l_1$, SSIM, and PSNR metrics were calculated by comparing the original Cityscapes image (as the ground truth) with the inpainted image.} To evaluate changes in visual attention, we compare different methods in terms of the average $v_o$ and $v_d$ (as defined in Section \ref{sec:3.4}).

\textbf{Comparison methods.} {Five state-of-the-art image-painting methods with open source codes were compared with the proposed scheme, including PICNet~\cite{zheng2019pluralistic}, RFR-Net\cite{li2020recurrent}, CR-Fill~\cite{zeng2021cr}, MST-net~\cite{cao2021learning}, and the original LaMa model~\cite{suvorov2022resolution}. PICNet, CR-Fill, MST-net and the LaMa model were trained on the large-scale Places2~\cite{zhou2017places} dataset that provides 8 million training images and contains highly diverse urban and street scenes. The RFR-Net was trained on the Paris StreetView~\cite{doersch2012makes} dataset (with 15,000 images). Note that PICNet, RFR-Net, and MST-net model can only handle squared images and require an input image with a fixed size of $256\times256$. We used the pre-trained models of these methods in our experiment.}

\subsection{Comparison to State-of-the-art Inpainting Methods}\label{sec:4.2}
\textbf{Quantitative comparison.} For a fair comparison, we input the same inpainting masks based on the proposed hierarchical salient object detection to all methods. Table \ref{tab:1} shows the quantitative results of inpainting performance in five metrics. {Benefiting from the FCCs for a high-receptive field in the inpainting network, the LaMa model outperforms other baselines when considering all metrics without the fine-tuning on the Cityscapes dataset. Our framework fine-tuned the model on the Cityscapes dataset (with 18,000 images), leading to the best performance in most metrics.} 

\begin{table}[t]
\footnotesize
\caption{Quantitative evaluation of inpainting performance on Cityscapes dataset}
\centering
\begin{tabular}{c|c|c|c|c|c} 
\hline
Method & SSIM$\uparrow$ & PSNR$\uparrow$   &$l_1\downarrow$ & LPIPS$\downarrow$  & FID$\downarrow$  \\ 
\hline
PICNet~\cite{zheng2019pluralistic}  & 0.8847 & 25.77&  0.0189  & 0.8213  &  5.08\\ 
\hline
RFR-Net~\cite{li2020recurrent}  & 0.9089  & 25.00 &   0.0158 &\textbf{0.7108}  & 5.21\\ 
\hline
CR-Fill~\cite{zeng2021cr}   & 0.9134 & 25.65 & 0.0171 & 0.7236 &5.08 \\ 
\hline
MST-net~\cite{cao2021learning}  & 0.8416  & 23.79 & 0.0309  & 1.1859  & 5.29  \\ 
\hline
{LaMa\cite{suvorov2022resolution}}   & {0.9118} & {25.88} &{0.0171}& {0.7261} & {5.07} \\
\hline
Ours    & \textbf{0.9175} & \textbf{26.38}  &  \textbf{0.0154}  &  0.7296 &  \textbf{5.04}  \\
\hline
\end{tabular}
\label{tab:1}
\end{table}

\textbf{Qualitative comparison.} Figure \ref{fig:3} compares the visual quality of different image-inpainting methods in street view images. We can see that the generated domain of our method is semantically more coherent with the surrounding regions, with fewer visual artifacts. Moreover, unlike RFR-Net, which has high performance in quantitative evaluation but tends to have poor inpainting performance for distracting objects in edge areas, the results of our method are more visually natural.

\subsection{Comparison of Different Inpainting Masks}\label{sec:4.3} 
\textbf{Quantitative comparison.} To show the effectiveness of the proposed hierarchical salient object selection module in generating inpainting masks, we compare different mask schemes, including only $\textbf{O}_1$ semantic level (human), only $\textbf{O}_2$ semantic level (Vehicle), only $\textbf{O}_3$ semantic level (Sign), all categories, and our hierarchical selection method. Table \ref{tab:2} presents the comparison results in terms of both inpainting performance and changes in visual attention. From the five inpainting performance metrics, we can see that only inpainting the semantic region $\textbf{O}_1$ with Human achieves the best performance. This is reasonable since the $\textbf{O}_1$ region with the Human tends to be much smaller than other objects in street view images, therefore resulting in a smaller inpainting domain and better image quality. However, this also leads to the smallest changes in visual attention. In contrast, only inpainting the $\textbf{O}_2$ can effectively reduce visual distraction, but it still works worse in redirecting visual attention to the object of interest. Our hierarchical strategy achieves high values in both $v_o$ and $v_d$, and also performs better in terms of inpainting results than using all categories as masks.

\begin{figure}[t]
  \centering
  \includegraphics[width=1\linewidth]{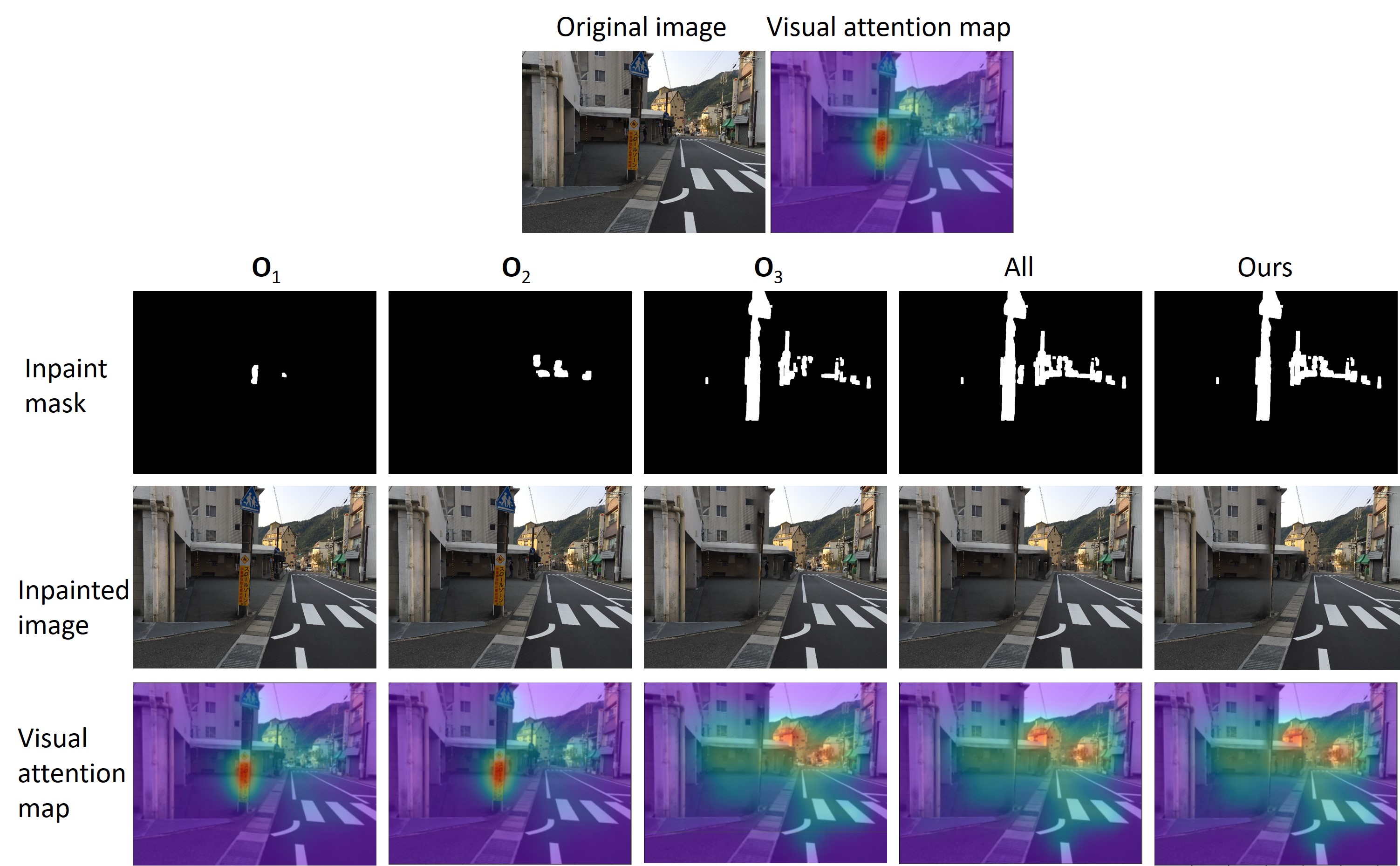}
  \caption{Visual performance of different inpainting masks on Cityscapes dataset. $\textbf{O}_1$ denotes Human semantic level, $\textbf{O}_2$ denotes Vehicle semantic level, and $\textbf{O}_3$ denotes Sign semantic level. Best view in color and zoom in.}
  \label{fig:4}
\end{figure}

\begin{table*}[t]
\scriptsize
\caption{Evaluating the performance of different inpainting masks}
\centering
\begin{tabular}{c|c|c|c|c|c|c|c} 
\hline
\multirow{2}{*}{Inpainted domain}  & \multirow{2}{*}{SSIM$\uparrow$} & \multirow{2}{*}{PSNR$\uparrow$} &\multirow{2}{*}{$l_1\downarrow$} & \multirow{2}{*}{LPIPS$\downarrow$}  & \multirow{2}{*}{FID$\downarrow$}  & \multicolumn{2}{c}{Visual attention change}     \\ 
\cline{7-8}
          &  &     &       &           &       &  $v_o$($\textbf{O}_o$ increase)$\uparrow$  &$v_d$ ($\textbf{O}_d$ reduction)$\uparrow$ \\ 
\hline
$\textbf{O}_1$   &   0.9923  &28.62   & 0.0019 & 0.1015  & 4.83 &  0.0533 &  0.0018       \\ 
\hline
$\textbf{O}_2$   & 0.9574  & 27.26  & 0.0088&  0.3804 & 5.01 &  0.2653 & 0.1469      \\ 
\hline
$\textbf{O}_3$  & 0.9606  & 27.88    & 0.0067&  0.4165  & 4.70  & 0.3113  & 0.0361     \\ 
\hline
All categories  &  0.9164  & 26.14 & 0.0155  &0.7298 & 5.05  &  1.1155  &  0.1226      \\ 
\hline
Ours   & {0.9175} & {26.38} & {0.0154}  &  0.7296 & 5.04    &  1.0483 & 0.1403     \\
\hline
\end{tabular}
\label{tab:2}
\end{table*}
\textbf{Qualitative comparison.}
We show the visual results of different mask generation schemes in Figure \ref{fig:4}. Similar to the results in Table \ref{tab:2}, small mask regions lead to high-quality images but small gaze changes. Our method can well balance the two requirements based on hierarchical salient object selection.

\subsection{Comparison of Different Urban Morphologies}\label{sec:4.4}
\textbf{Quantitative comparisons.} Due to the diversity of street view images in location and acquisition, the need and performance of saliency-guided image inpainting will differ greatly. For example, street view images captured in bustling cities will be more widely used for last-meters wayfinding and contain more distracting objects than those captured in $motorway$. To explore how the urban morphology in street view images will influence our method, we used the urban canyon classification method~\cite{hu2020classification} to classify the street scenes in Cityscapes dataset into four categories based on the street aspect ratio $\alpha$\footnote{$\alpha$ is the ratio of the canyon height to the canyon width.}. Note that images with a higher $\alpha$ tend to be captured in the downtown area, while images with a smaller $\alpha$ may contain suburb scenes. From the comparison results in Table \ref{tab:3}, we can observe that a smaller aspect ratio $\alpha$ leads to better inpainting performance. This is reasonable since street view images captured in the suburb tend to contain less distracting objects, meaning smaller inpainting domains. Meanwhile, images of non-street canyons with $\alpha$ as 0 may contain no or fewer location-related objects. As a result, our method obtains the largest $v_o$ but the smallest $v_d$ distracting objects. For street scenes with a large ratio $\alpha$, our method shows a greater advantage in redirecting visual attention to objects of interest.

\begin{figure}[t]
  \centering
  \includegraphics[width=1\linewidth]{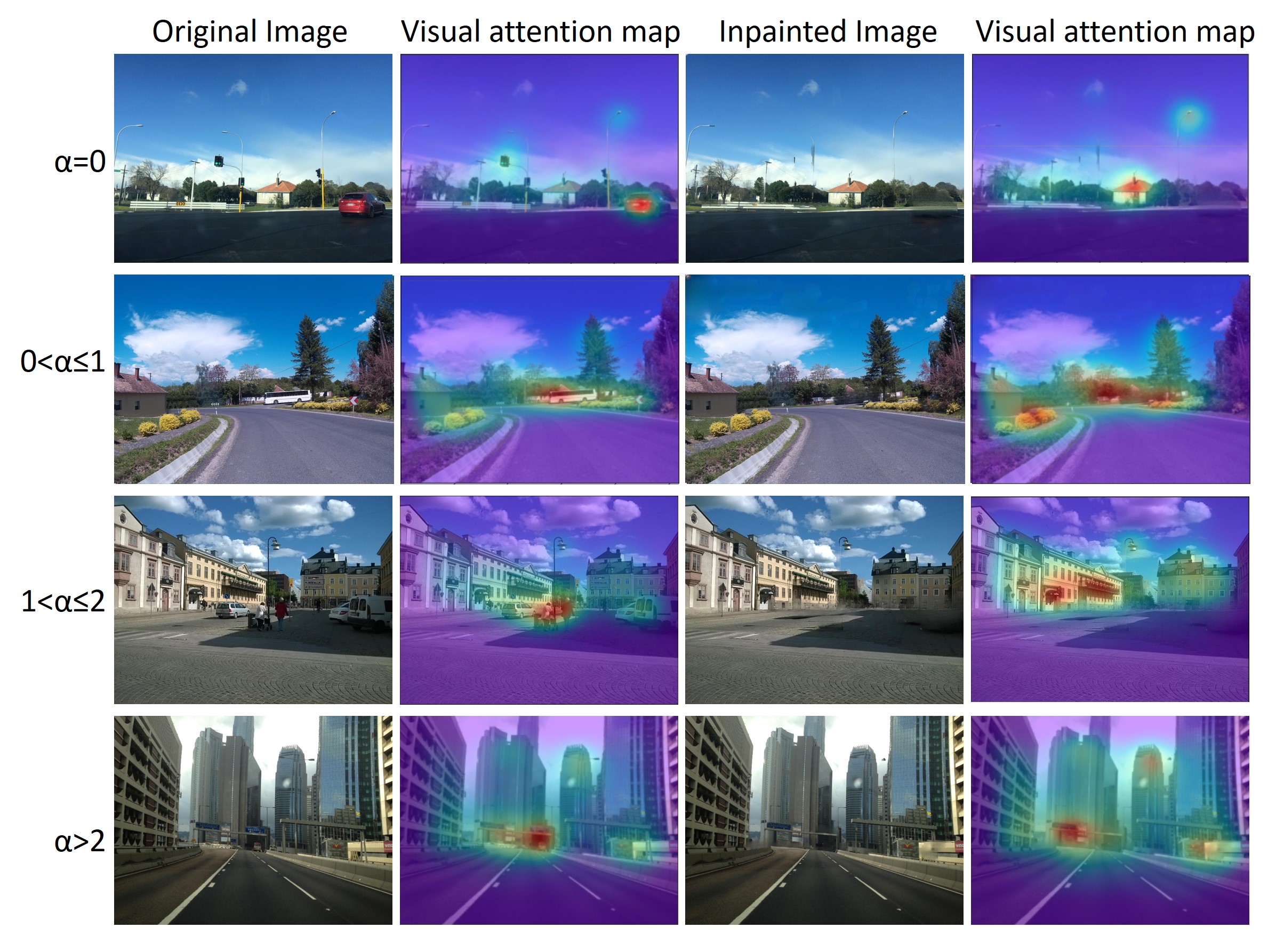}
\vspace{-0.85cm}
  \caption{Visual performance of our proposed method in different types of street view images. Best view in color and zoom in.}
  \label{fig:5}
\end{figure}

\begin{table*}[t]
\scriptsize
\caption{Quantitative evaluation of the influence of $\alpha$ on the performance of the inpainting.}
\centering
\begin{tabular}{c|c|c|c|c|c|c} 
\hline
\multirow{2}{*}{Street aspect ratio} &\multirow{2}{*}{$l_1\downarrow$} & \multirow{2}{*}{SSIM$\uparrow$} & \multirow{2}{*}{PSNR$\uparrow$}  & \multirow{2}{*}{LPIPS$\downarrow$}    & \multicolumn{2}{c}{Visual attention change}     \\ 
\cline{6-7}
          &  &     &       &              &  $v_o$($\textbf{O}_o$ increase)$\uparrow$ &$v_d$ ($\textbf{O}_d$ reduction)$\uparrow$ \\
\hline
$\alpha=0$      &  0.0130 & 0.9282 & 27.43 & 0.6886  & 1.1910 & 0.0732  \\ 
\hline
$0<\alpha \leq1$   & 0.0175 & 0.9095 & 25.36 & 0.7595    & 0.8618  & 0.1903  \\ 
\hline
$1<\alpha \leq2$   &  0.0180  & 0.9046  &25.18  & 0.7871   &  0.9341 & 0.2326  \\ 
\hline
$\alpha>2$    & 0.0210  & 0.8923  & 24.29  &0.8172     &  1.1441 & 0.2318     \\
\hline
\end{tabular}
\label{tab:3}
\end{table*}

\textbf{Qualitative comparisons.}
Figure \ref{fig:5} illustrates the visual results of the proposed method in different types of street view images. For images captured in non-canyon areas ($\alpha$ = 0), buildings tend to be more isolated and easier to focus on after removing distracting objects and therefore lead to higher $v_o$ (as obtained in Table \ref{tab:3}). In contrast, our method shows more promising performance for street view images with larger street aspect ratios. Human gaze can be effectively redirected to location-related landmarks; meanwhile, inpainted images do not have obvious visual artifacts. 

\begin{figure}[!h]
\centering
\subfigure[\scriptsize{last-meters wayfinding based on original image}]{
\begin{minipage}[t]{0.5\linewidth}
\centering
\includegraphics[width=2.8in]{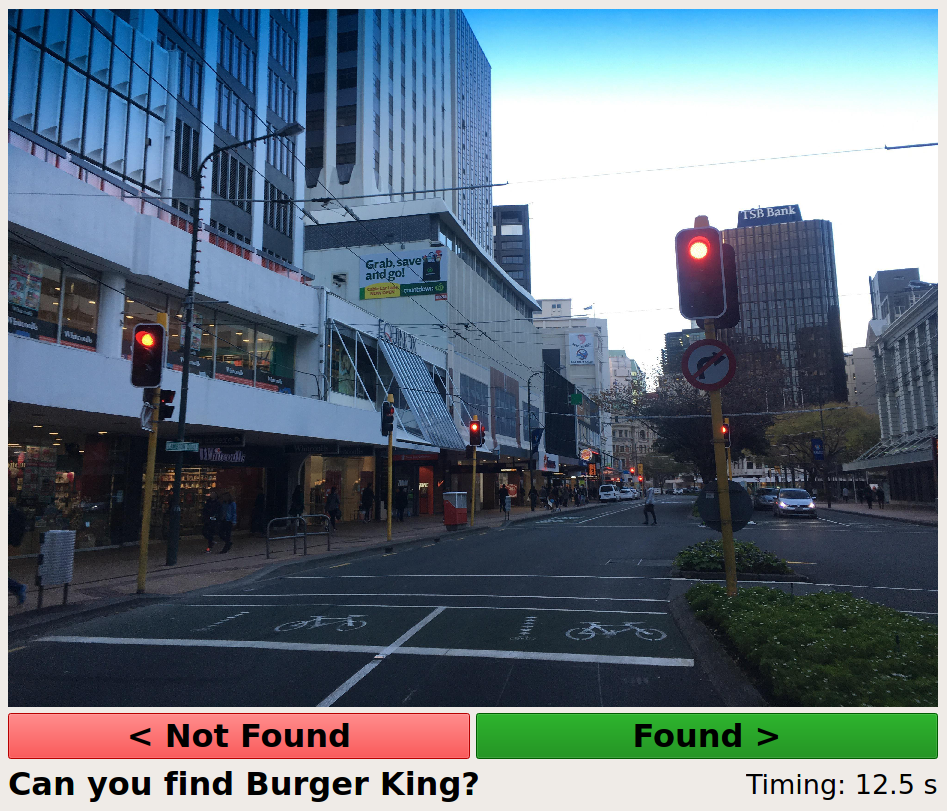}
\end{minipage}%
}%
\subfigure[\scriptsize{last-meters wayfinding based on inpainted image}]{
\begin{minipage}[t]{0.5\linewidth}
\centering
\includegraphics[width=2.8in]{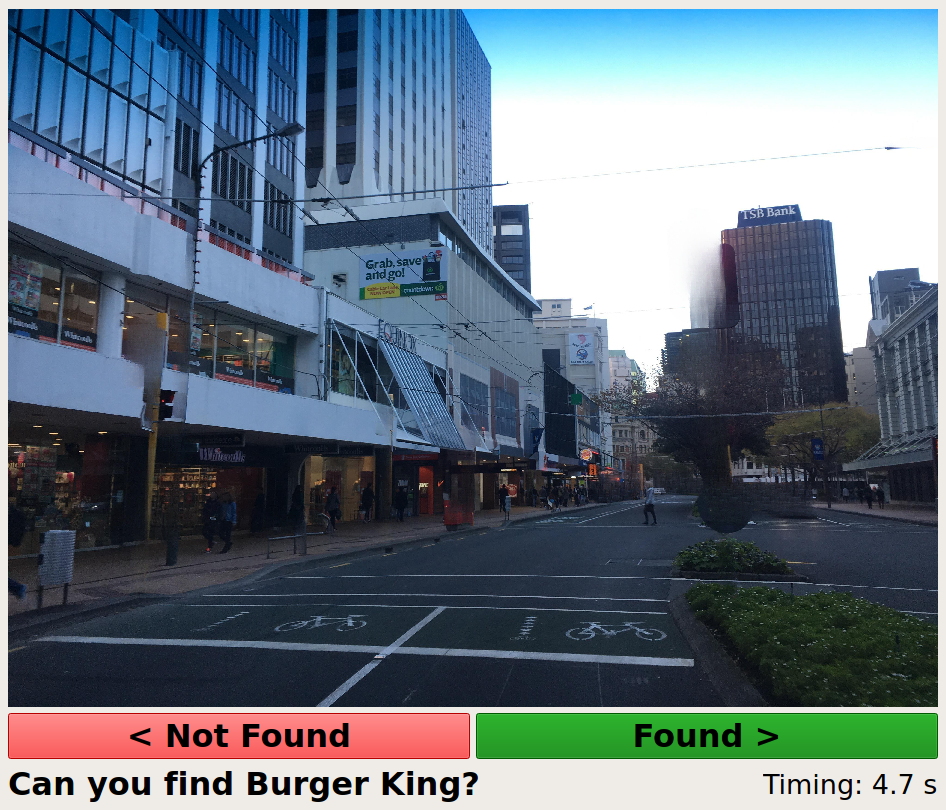}
\end{minipage}%
}%
\vspace{-0.15cm}
\subfigure[\scriptsize{last-meters wayfinding based on original image}]{
\begin{minipage}[t]{0.5\linewidth}
\centering
\includegraphics[width=2.8in]{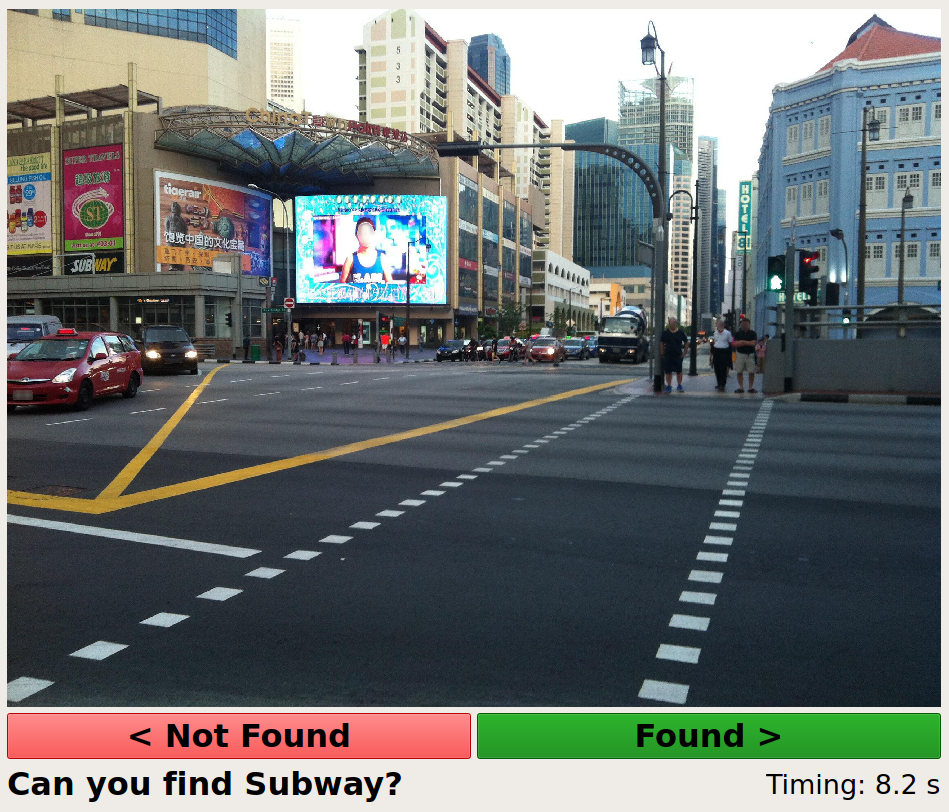}
\end{minipage}
}%
\subfigure[\scriptsize{last-meters wayfinding based on inpainted image}]{
\begin{minipage}[t]{0.5\linewidth}
\centering
\includegraphics[width=2.8in]{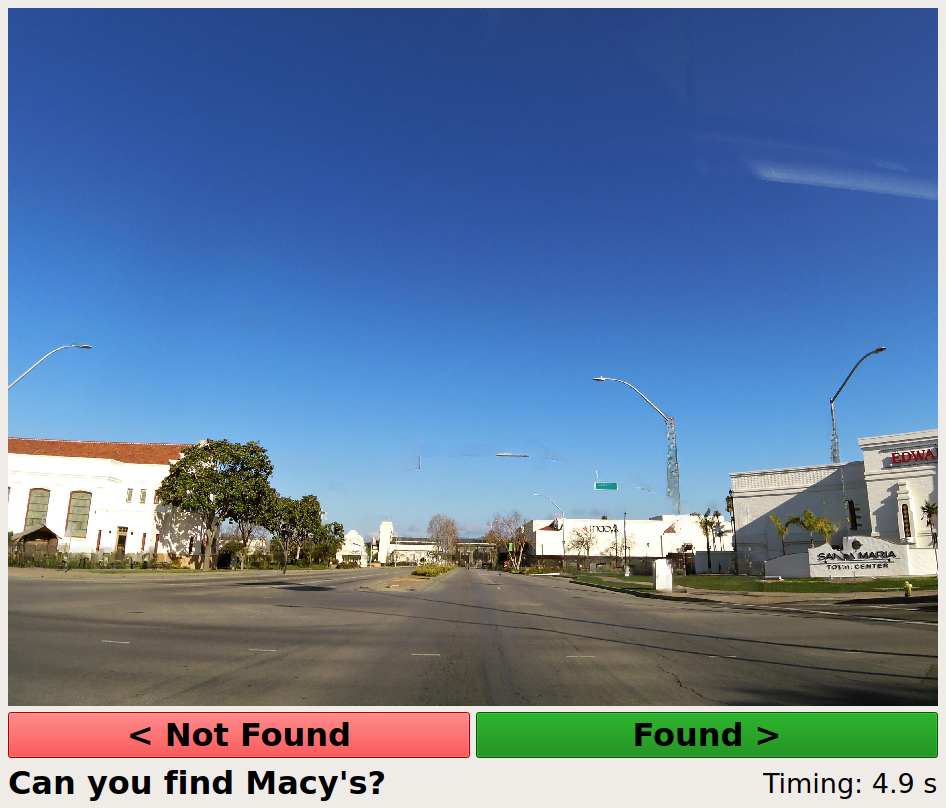}
\end{minipage}
}%
\centering
\vspace{-0.25cm}
\caption{Self-developed program for human last-meters wayfinding evaluation. Paired images, for example, examples in (a) and (b), are used to report results for comparison, while unpaired images, e.g., examples in (c) and (d), are shown to each volunteer to ensure that the finding scene is not repeated.}
\label{fig:9}
\end{figure}

\subsection{Comparison of Last-Few-Meters Wayfinding Efficiency}\label{sec:4.5}
\label{sec:4.5}
To show the effectiveness of our method in practical last-meters wayfinding applications, we conducted a controlled human-based evaluation to test the last-meters wayfinding efficiency before and after applying our inpainting method to street-view images. We carefully selected 50 original street view images with location-based buildings (e.g., landmarks, shopping malls, stores, and supermarkets) and then obtained their inpainted images based on our method. The 50 pairs of images were randomly divided into two subsets. Data\_1 with 25 pairs and Data\_2 with 25 pairs. Totally 10 volunteers (5 men and 5 women, ages 20 to 55) were asked to find a specific location using our self-developed platform (as shown in Figure \ref{fig:9}). Volunteers were randomly divided into two groups: Group\_1 with 5 persons and Group\_2 with 5 persons, and conducted the evaluation according to the settings in Table \ref{tab:4a}. The search time between the image display and the ``Found" confirmation click was recorded. To reduce individual differences in searching and matching, we normalized the time values of each volunteer between 0 and 1. Table \ref{tab:4b} shows the average value of normalized search time, and we can see that our saliency-guided inpainting method can reduce approximately 38\% search time by removing distracting objects from street view images.

\begin{table}[!t]
\footnotesize
\begin{floatrow}
\capbtabbox{
\begin{tabular}{c|c|c} 
\hline
Subset & Group\_1 & Group\_2 \\
\hline
Data\_1&  Original & Inpainted  \\ 
\hline
Data\_2 &  Inpainted  & Original  \\ 
\hline
 \end{tabular}
}{
 \caption{Human-based evaluation settings in data and volunteers}
 \label{tab:4a}
}
\capbtabbox{
\begin{tabular}{c|c|c|c} 
\hline
Subset & Original & Inpainted  & Improvement \\
\hline
Data\_1&  0.4165 & 0.2547  & 38.85\%\\ 
\hline
Data\_2 & 0.4569 &  0.2834 &  37.97\%\\ 
\hline

 \end{tabular}
\label{tab:4}
}{
 \caption{Comparison of human Last-Few-Meters wayfinding efficiency in normalized search time}
 \label{tab:4b}
}
\end{floatrow}
\end{table}

\subsection{{Influence of different components}}
\begin{figure}[!h]
  \centering
  \includegraphics [width=1\linewidth]{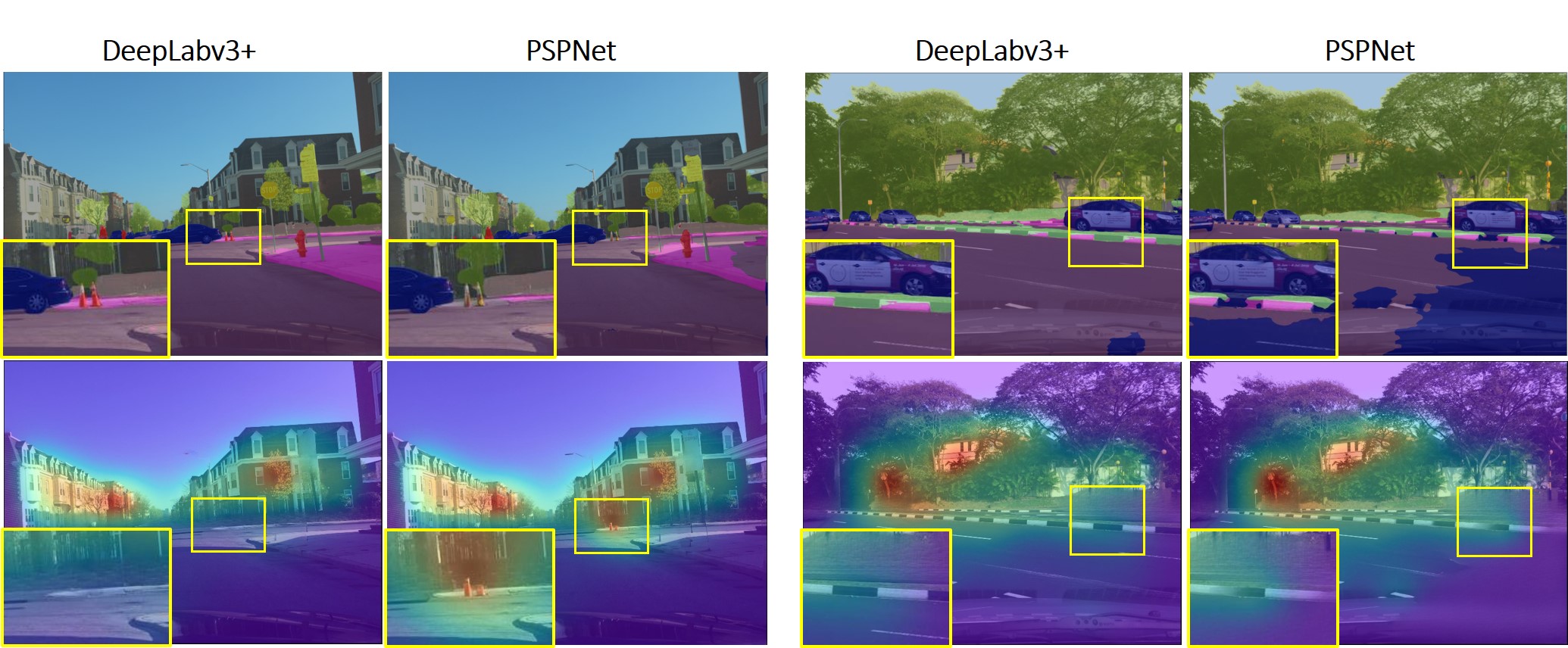}
  \caption{{Influence of semantic segmentation performance on image inpainting and visual attention changes in wayfinding. Images in the first row are the semantic segmentation results, and images in the second row are the human visual attention maps on the inpainted images.}}
  \label{fig:seg}
\end{figure}

{To explore the sensitivity of the proposed framework to different components, including semantic segmentation, salient object detection, and human visual attention detection, we performed comparison experiments to demonstrate the influence of these components on the performance of last-way finding. For the semantic segmentation module, we compared DeepLabv3+ \cite{chen2018encoder} used in our framework with the PSPNet \cite{zhao2017pyramid} model, which has been proved to perform worse than the DeepLabv3+ model on several datasets~\cite{chen2018encoder}, including the Cityscapes data~\cite{imam2019semantic}. Other modules were set the same as our proposed framework, i.e., with the $U^2Net$ model \cite{qin2020u2} for salient object detection and UNISAL \cite{drostejiao2020} for visual attention detection. The results in the first three rows of Table \ref{tab:9} show that the PSPNet \cite{zhao2017pyramid} with worse segmentation performance achieves smaller visual attention changes than the DeepLabv3+ model. This means that the accuracy of semantic segmentation can obviously influence the final wayfinding efficiency. The more accurate the segmentation model is, the better the inpaiting method will perform, and therefore the larger improvement in last-meters wayfinding efficiency will be achieved. 
In Figure \ref{fig:seg}, we further present two testing examples to show the influence of segmentation performance on image inpainting and visual attention changes in wayfinding. We can also see that the worse performance of PSPNet leads to less visual distraction reduction for wayfinding than the DeepLabv3+ model.}
\begin{table*}[t]
\footnotesize
\caption{{Evaluation of the influence of different model components}}
\newcommand{\tabincell}[2]{\begin{tabular}{@{}#1@{}}#2\end{tabular}} %
\centering
\begin{tabular}{c|c|c|c} 
\hline
\multirow{2}{*}{Module}      & \multirow{2}{*}{Method} & \multicolumn{2}{c}{Visual attention change} \\ 
\cline{3-4}   &       & $v_o$($\textbf{O}_o$ increase)$\uparrow$  &$v_d$ ($\textbf{O}_d$ reduction)$\uparrow$          \\ 
\hline
\multirow{2}{*}{Segmentation}   & {PSPNet~\cite{zhao2017pyramid}}  &  {0.9189}        & {0.0450}                          \\ 
\cline{2-4}   & DeepLabv3+~\cite{chen2018encoder} &  1.0483 &  0.1403   \\  
\hline
\multirow{2}{*}{\tabincell{c}{ Salient object\\detection} }  & {BASNet \cite{qin2019basnet}} &  {1.0762}        &  {0.1362}       \\ 
\cline{2-4}   &$U^2Net$ model \cite{qin2020u2} &  1.0483 &  0.1403   \\ 
\hline
\multirow{2}{*}{\tabincell{c}{ Visual attention\\detection}} & {TranSalNet \cite{lou2022transalnet}} &  {1.1112}        &   {0.1205}                                                 \\ 
\cline{2-4}  &UNISAL~\cite{drostejiao2020}  &   1.0483 &  0.1403                 \\
\hline
\end{tabular}
\label{tab:9}
\end{table*}

{Although the two saliency detection models, i.e., $U^2$Net \cite{qin2020u2} and UNISAL \cite{drostejiao2020} are not specifically trained for street view images, both models are trained with a high diversity of images that includes various street view images. It is reasonable to believe that these two models can well extract the salient region and human attention in street view images on the Cityscapes dataset due to the overlapped image content with their training data. To further evaluate how saliency detection methods will influence the analysis, we compared the performance of different saliency detection models and human visual attention detection models. Specifically, for salient object detection, we compared $U^2$ Net \cite{qin2020u2} with the widely used BASNet \cite{qin2019basnet} and set other modules the same as our proposed framework. For human visual attention detection, we compared UNISAL \cite{drostejiao2020} with the recent TranSalNet \cite{lou2022transalnet} for human attention detection. The results in Table \ref{tab:9} show that the influence of these two modules on human attention changes in last-way finding is not significant, since they achieved quite close $v_o$ and $v_d$ values. Not only does it show the reliability of $U^2$Net and UNISAL used in our framework, but it also demonstrates the flexibility of our framework, which can be easily adjusted to other saliency detection methods.}

\subsection{Extended Application to Indoor Scenarios}

\begin{figure}[t]
  \centering
  \includegraphics[width=\linewidth]{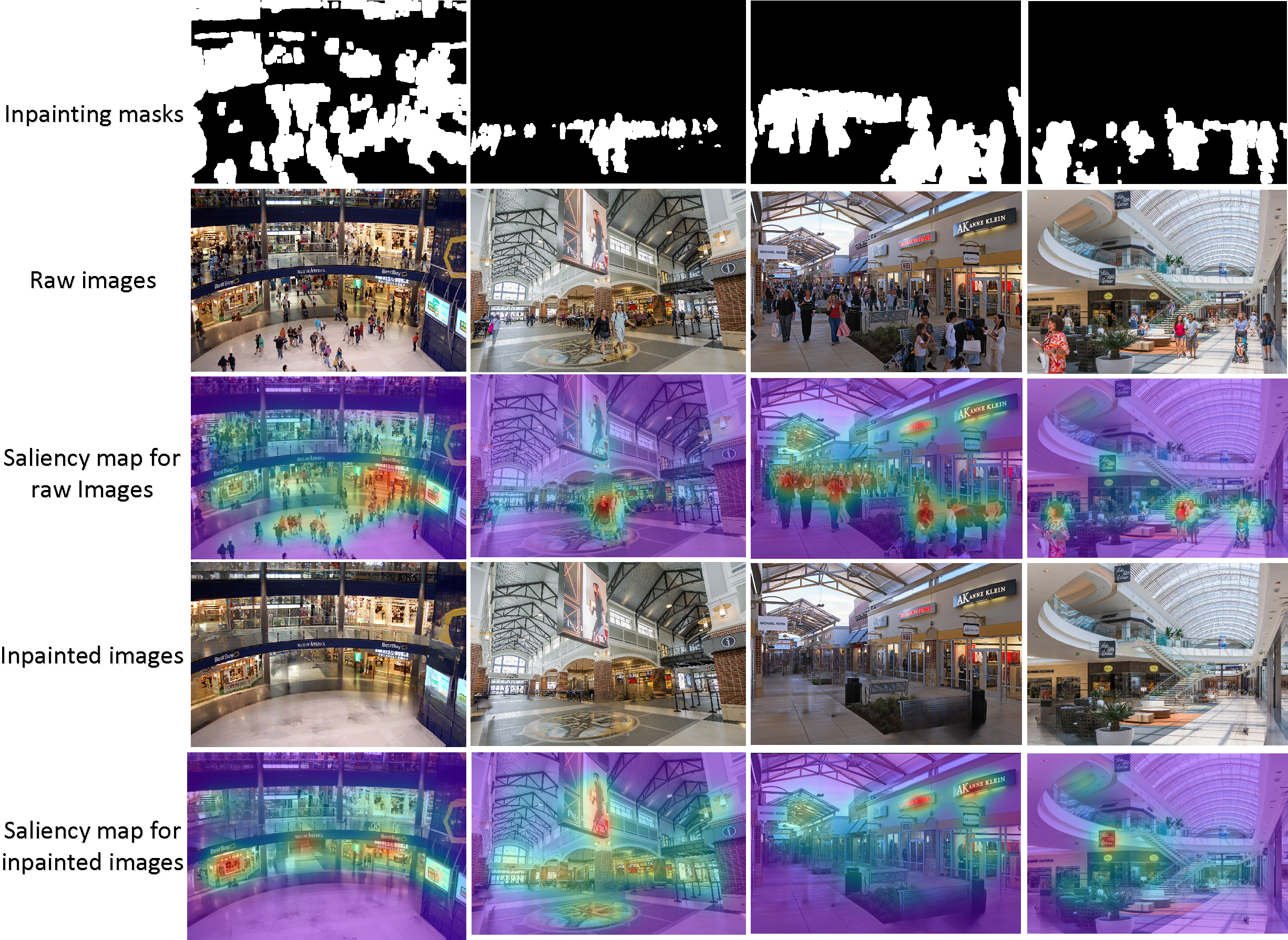}
  \vspace{-0.35cm}
  \caption{Examples of the proposed method in indoor scene application. Best view when zooming in.}
  \label{fig:10}
\end{figure}

Previous experimental results have demonstrated the effectiveness of the proposed method in improving the outdoor last-meters wayfinding. To illustrate the extensibility of our method, we show its application in indoor scenarios, where a particular brand store in a crowded mall may be difficult to identify in indoor navigation systems. To simulate this scenario, 2,000 images with interior scenes of shopping malls have been collected using the Bing Image Scraper\footnote{\href{https://pypi.org/project/bing-image-downloader/}{\url{https://pypi.org/project/bing-image-downloader/.}}}, and then the brand stores that must be found in each image are annotated. The qualitative results of our method in terms of the effects of inpainting and changes in human vision are shown in Fig. \ref{fig:10}. It can be observed that the dense and distracting crowd can be effectively removed to redirect human visual attention to store localization. 
Similarly to the settings in Table \ref{tab:4b}, the quantitative results of the human search time based on 50 pairs of indoor images are shown in Table \ref{tab:11}. We can see that our saliency-guided inpainting method can reduce approximately 28\% search time by removing distracting objects in indoor scene images. The reasons why the improvement of indoor scenarios is smaller than that of outdoor images (see the results in Table \ref{tab:4b}) can be attributed to the fact that the scene diversity of indoor images is lower than outdoor street-view images, including lighting condition, image quality, and distracting region size.

\begin{table}[]
\footnotesize
\begin{tabular}{l|c|c|c}
\hline
Subset  & Original & Inpainted & Improvement \\
\hline
Data\_1 & 0.5059   & 0.3707    & 26.72\%     \\
\hline
Data\_2 & 0.4669   & 0.3269    & 29.99\%     \\
\hline
\end{tabular}
\caption{Comparison of human Last-Few-Meters
wayfinding efficiency for an indoor scenario in normalized search time}
\label{tab:11}
\end{table}

\section{Discussions}\label{sec:5}  
\begin{figure}[t]
  \centering
  \includegraphics[width=1.0\linewidth]{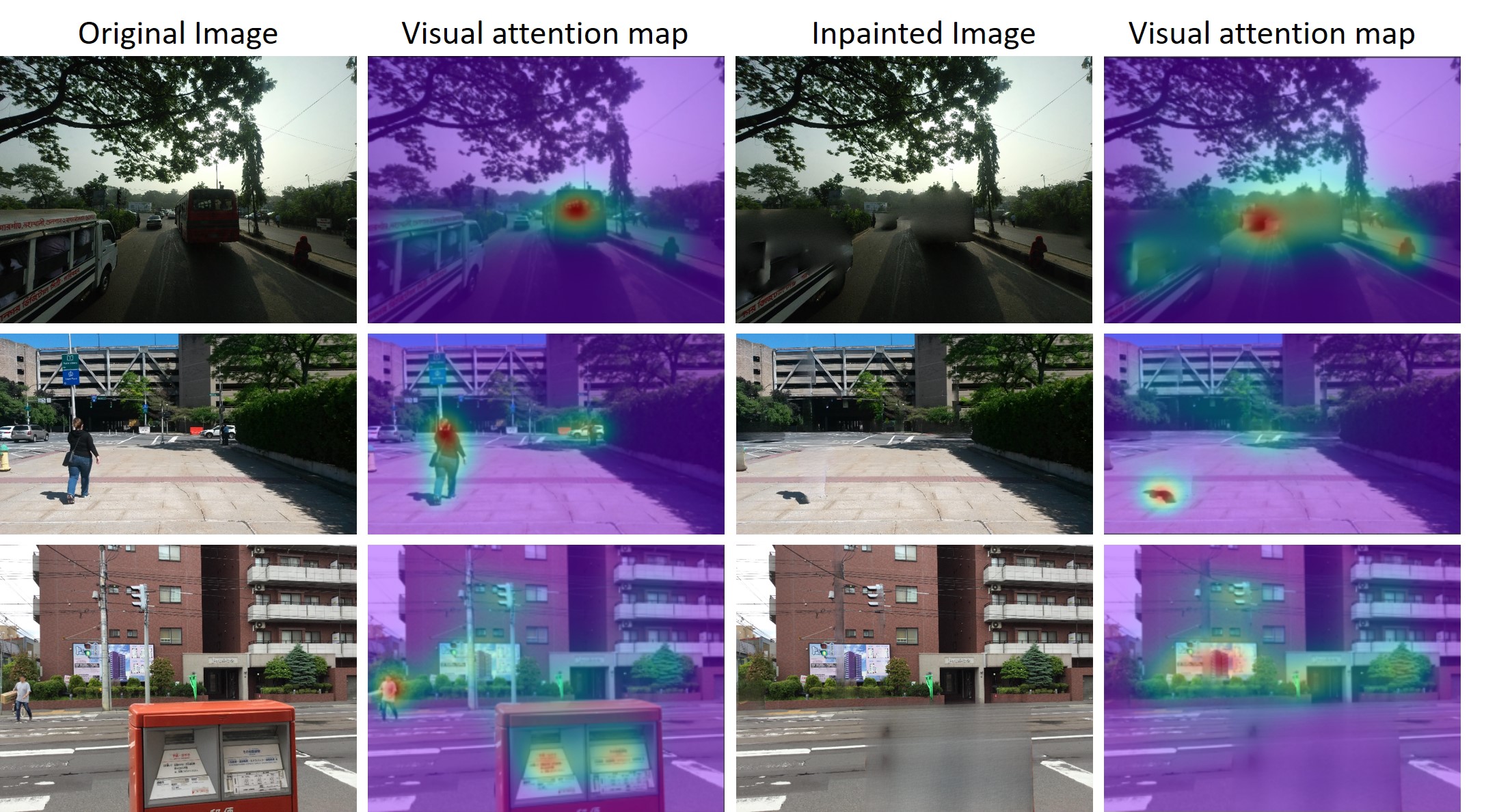}
  \caption{Examples with obvious visual inpainting artifacts in street view images under dark conditions (the first row), with shadows (the second row), and with large distracting object regions (the third row). Best view when zooming in.}
  \label{fig:6}
\end{figure}

The proposed saliency-guided image inpainting method has demonstrated promising performance in redirecting human visual attention from distracting objects to location-related objects in street view images. As the first study to integrate saliency detection into image inpainting for human last-meters wayfinding, there is much room for improvement of the current method. For example, inpainting performance can be further improved due to lingering visual artifacts in some challenging cases, as shown in Figure \ref{fig:6}. These artifacts can be attributed to the following three reasons. First, street-view images have high diversity, such as lighting conditions and acquisition location. Second, the semantic image segmentation method has limitations in detecting objects with complex structures, especially those with shadows. Third, the image-inpainting method shows unstable performance in removing large domains. 

Furthermore, the saliency-guided image inpainting in Figure \ref{fig:2} can be further optimized by end-to-end training. In the current implementation of our framework, the saliency detection and image-inpainting steps remain separate. It is possible to construct an objective function (e.g., $v_o+v_d$) related to the visual attention of the object of interest. The goal of removing distracting objects can be formulated as a constrained optimization problem, such as maximizing the objective function with limited resources (e.g., running time) to support real-world applications \cite{bescos2019empty}. From the perspective of network design, end-to-end training of joint dynamic object detection and inpainting could further improve the performance of our model. We can start from the simulated street view images, such as the CARLA dataset \cite{Dosovitskiy17} and fine-tune the trained model on real-world street view images. 

The other interesting application is the street view image processing for privacy protection \cite{frome2009large,flores2010removing}. Instead of blurring faces or license plates, the sensitive information can be removed or replaced by virtual placeholders (e.g., faces and vehicles that do not exist). As the line between the physical world and the metaverse become fuzzy, we believe that this line of research is also applicable to privacy research, augmented reality, and intelligent transportation systems. Finally, given the rich redundancy in the temporal domain, the extension of this work to street-view video inpainting and depth-assisted image inpainting will remain as our future work.

\section{Conclusions}\label{sec:6}
Image-based last-meters wayfinding is easy to distract human visual attention due to the diverse and rich content in the street view images. In this paper, we make the first attempt to tackle this problem by designing a saliency-guided image inpainting framework, which aims to inpaint distracting objects on a semantic level to redirect human visual attention to location-related landmarks. Our method designs a hierarchical salient object selection module to generate inpainting masks and removes distracting objects based on the large-mask inpainting method with FFCs. Experimental results with qualitative and quantitative evaluations of street view images have validated the effectiveness of the proposed method. We believe that our framework will not only benefit image-based last-meters wayfinding but also can extend to other types of image restoration.

\section*{ACKNOWLEDGEMENTS}\label{ACKNOWLEDGEMENTS}

This work is partially supported by the DOJ under grant NIJ 2018-75-CX-0032, NSF under grant OAC-1839909, IIS-1951504 and the WV Higher Education Policy Commission Grant (HEPC.dsr.18.5).

 \bibliographystyle{elsarticle-num} 
 \bibliography{cas-refs}





\end{document}